\documentclass[a4paper,conference]{IEEEtran}

\ifCLASSINFOpdf
  \usepackage[pdftex]{graphicx}
\else
\fi
\usepackage{dsfont}
\usepackage{amsmath}
\usepackage{amssymb,amsfonts,bbm,euscript}       
\usepackage{mathabx}

\newcommand{\Exp}{\mathsf{E}}

\newcommand{\f}{\mathsf{f}}
\newcommand{\g}{\mathsf{g}}
\newcommand{\h}{\mathsf{h}}

\newcommand{\G}{\mathsf{G}}
\newcommand{\T}{\mathsf{T}}
\newcommand{\C}{\mathsf{C}}
\newcommand{\I}{\mathsf{I}}

\newcommand{\cC}{\EuScript{C}}

\begin{document}
%
\title{Image De-Quantization Using Generative Models\\ as Priors}

\author{\IEEEauthorblockN{Kalliopi Basioti}
\IEEEauthorblockA{Department of Computer Science\\
Rutgers University\\
New Brunswick, New Jersey, USA\\
Email: kib21@scarletmail.rutgers.edu}
\and
\IEEEauthorblockN{George Moustakides}
\IEEEauthorblockA{Department of Electrical and Computer Engineering\\
University of Patras\\
Patras, Greece\\
Email: moustaki@upatras.gr}}

\maketitle

\begin{abstract}
Image quantization is used in several applications aiming in reducing the number of available colors in an image and therefore its size. De-quantization is the task of reversing the quantization effect and recovering the original multi-chromatic level image. Existing techniques achieve de-quantization by imposing suitable constraints on the ideal image in order to make the recovery problem feasible since it is otherwise ill-posed. Our goal in this work is to develop a de-quantization mechanism through a rigorous mathematical analysis which is based on the classical statistical estimation theory. In this effort we incorporate generative modeling of the ideal image as a suitable prior information. The resulting technique is simple and capable of de-quantizing successfully images that have experienced severe quantization effects. Interestingly, our method can recover images even if the quantization process is not exactly known and contains unknown parameters.
\end{abstract}


%
\IEEEpeerreviewmaketitle

\section{Introduction}
%

Visual quality of images is largely affected by the diversity of its color palette, namely the number of available colors for its digital representation. In \textit{image quantization}, the size of the original color palette is reduced by substituting every color of the original image with the closest color in the reduced-size-palette. Image quantization occurs in many cases. Examples are photography with digital cameras that have limited color palettes; image compression where we decrease the color levels by reducing the number of bits per pixel to represent to corresponding color. This in turn results in smaller storage space requirement or processing with smaller bit accuracy. Although image quantization may affect the size of the color palette whenever it is deemed necessary, it is clear that it is a \textit{lossy procedure}. Recovering the lost information, namely the original chromatic representation, it is not straightforward nor is the realistic enrichment of the number of colors.

There are various image quantization techniques starting with the simplest where we uniformly quantize each chromatic component and ending with more sophisticated methods where one can, for example, employ the K-means procedure to cluster the pixels, with each pixel being represented by a 3-D vector containing the chromatic components, and then selecting a representative for each cluster \cite{pollard}. In what follows, for simplicity, we focus mostly on uniform quantization, but our results are readily extendable to more complicated quantization schemes.

In our present work we are interested in the problem of \textit{de-quantization}, namely, from a quantized image to recover the original multi-color-level version. It is obvious that the de-quantization problem is ill-posed since many different images after being quantized can yield the same quantized version. In order to be able to obtain a single solution it is therefore clear that we must impose meaningful constraints. These constraints can be vaguely distinguished into two major catergories. In the first we attempt with the constraints to capture the \textit{structural properties} of the image. Well known methods in this category include: a)~the total variation method \cite{afonso} where one assumes that natural images tend to be smooth with not very high-frequency components; b)~low rank methods \cite{hu} where the original image, when represented as a matrix, accepts a low rank decomposition; c)~sparse methods \cite{yang} where the original image enjoys a sparse representation in the space spanned by a suitable dictionary.

The second category involves constraints that attempt to capture the \textit{statistical behavior} of the original image. More precisely, the original image is assumed to be a realization of a \textit{random image} where for the latter we have available a statistical description in the form of a joint probability density of its elements (chromatic components of each pixels). Instead of the probability density one may also employ an equivalent statistical representation as, for example, a \textit{generative model}. By properly engaging this prior knowledge one may arrive in solving the de-quantization problem very efficiently. Currently, there exist very few methods that fall under the second category. Early such techniques were using the generative model only partially \cite{bora} by limiting it to the generator function. This idea frequently resulted in convergence to visually incorrect estimates. In order to improve upon the original estimates the discriminator function (coming from an adversarial design of the generator model) was also employed in \cite{yeh, yeh2} in the form of a regularizer term. However, this idea increases the computational complexity significantly since it requires off-line tuning of the weighting parameter of the regularizer term.

It is clear that the two categories rely on completely different forms of constraints with the first assuming some form of smoothness in the image or some transformed version of the image and the second relying on a \textit{frequentistic} description of the original image, namely, how \textit{probable} is the occurrence of the existing realization.

In this work we adopt the second category of constraints and assume that a generative model is available that captures the statistical behavior of the original image. We basically intend to specialize the results developed for image restoration \cite{basioti} to the de-quantization problem. This specialization is not straightforward since it demands proper mathematical analysis and the definition of suitable functions which are not mentioned in the general problem treated in \cite{basioti}. As in \cite{basioti}, we intend to reach our goal through a rigorous mathematical analysis which is based on the classical statistical estimation theory and which will lead us to a very well defined optimization problem, the solution of which will provide the desired estimate. Regarding the final optimization we must add that it will not contain any parameters that require fine-tuning using proper pre-processing, as is the case with the existing methods we mentioned before. But what is even more worth mentioning is the fact that we will be able to treat successfully problems which contain unknown parameters (e.g.~not exactly known quantization strategies), something that has no equivalent in existing methods. Next, let us briefly recall relevant results from statistical estimation.

\section{Statistical Estimation Theory}
Even though a more detailed version of the results we are going to use appears in \cite{basioti}, for completeness and self sufficiency of our article we present without major discussion theoretical elements that are relevant to the problem of interest. We begin by considering two random vectors $Y,Z$ where $Z$ is hidden while $Y$ is observed. Using $Y$, we would like to \textit{estimate} the hidden vector $Z$ when $Y,Z$ are statistically related, with their relationship captured by the joint probability density function $\f(Y,Z)$. 

Using $\f(Y,Z)$ one can apply the well known Bayesian estimation theory \cite[pp. 259–279]{veeravalli} in order to estimate $Z$ from $Y$. We recall that an estimator is any \textit{deterministic function} $\hat{Z}(Y)$ of $Y$ and it is by the application of the Bayesian methodology that we can identify the optimum estimator. The Bayesian approach requires the definition of a cost function $\C(\hat{Z},Z)$ which places a cost on each combination $\{\hat{Z},Z\}$ of estimate and true value. Optimum is considered the estimator that minimizes the average cost $\Exp[\C(\hat{Z}(Y),Z)]$, where expectation is with respect to $Z$ and $Y$.

There are various meaningful cost functions and their corresponding optimum estimators. We recall \cite[pp.~267-268]{veeravalli} the minimum mean square error (MMSE) criterion which leads to the conditional mean $\Exp[Z|Y]$ estimator; the minimum mean absolute error (MMAE) criterion which leads to the conditional median estimator \cite[page 267]{veeravalli} and finally the maximum a-posteriori probability (MAP) estimator which is defined as
\begin{multline}
\hat{Z}=\text{arg}\max_Z\f(Z|Y)\\
=\text{arg}\max_Z\frac{\f(Y,Z)}{\f(Y)}=\text{arg}\max_Z\f(Y,Z),
\label{eq:1.2}
\end{multline}
that proposes as optimum estimate the \textit{most likely} $Z$ given the observations $Y$. It is the MAP estimator we are going to adopt in our methodology since it has proven itself to yield efficient solutions in other image restoration problems \cite{basioti}. 

There is an additional reason we prefer the MAP estimator. In particular, it turns out that it also allows, without much extra complication, the treatment of estimation problems containing unknown parameters. As we argue in \cite{basioti}, if the joint density of $Y$ and $Z$ is of the form $\f(Y,Z|\gamma)$ containing a vector $\gamma$ of unknown parameters then, following again the Bayesian approach and assuming the most uninformative prior density for $\gamma$, in the form of an \textit{improper uniform}, we arrive at the following extension of the MAP estimator in the presence of unknown parameters
\begin{equation}
\hat{Z}=\text{arg}\max_Z\left\{\max_{\gamma\in\cC}\f(Y,Z|\gamma)\right\},
\label{eq:optim}
\end{equation}
where $\cC$ some known set. As we can see we first perform a \textit{maximum likelihood} estimate of the parameter vector $\gamma$ and then we MAP-optimize over the desired $Z$. We would like to stress that the optimization proposed in \eqref{eq:optim} basically assumes that the parameters $\gamma$ are \textit{realization dependent}. In other words if $Y$ denotes our quantized image then $\gamma$ may be different for each image we are asked to process. This assumption clearly excludes methods where one may use past data to learn $\gamma$ before hand and then treat the joint density of $Y$ and $Z$ as completely known. 

\section{Generative Models and Parametric Transformations}\label{sec:3}
A generative model is comprised of the generator function $X=\G(Z)$ and the input distribution $Z\sim\h(Z)$. Regarding the problem of interest it is completely unimportant how the generative model $\{\G(Z),\h(Z)\}$ is obtained. We can use adversarial approaches \cite{goodfellow, karras, basioti2}, non-adversarial approaches \cite{basioti3, dziugaite} or VAEs \cite{doersch}. Each one of these methods can provide the necessary pair $\{\G(Z),\h(Z)\}$ which is needed for our analysis. We would like to stress that we do not require any discriminator function as existing techniques since this function might not even exist if we do not train our generator using adversarial methods.

From the above it is clear that we are interested in a random vector $X$. Instead of describing its statistical behavior through the corresponding probability density $\f(X)$ we assume that $X$ is the output of the \textit{known} deterministic generator function $X=\G(Z)$ with the input $Z$ being random and following the \textit{known} probability density $\h(Z)$.

The vector $X$ denotes the ideal image and we subject $X$ to a deformation through a \textit{deterministic function} $\T(X,\alpha)$ where $\T(X,\alpha)$ has known functional form and a number of unknown parameters expressed with the vector $\alpha$. Finally to this deformed outcome we add noise $W$ which leads to the observed vector $Y$. The noise vector $W$ follows a probability density $\g(W|\beta)$ which has known functional form but may contain a number of unknown parameters expressed with the vector $\beta$. Summarizing, if $Y$ is our observation vector then
\begin{equation}
Y=\T(X,\alpha)+W=\T\big(\G(Z),\alpha\big)+W.
\end{equation}
Regarding our problem of interest $Y$ corresponds to the quantized image and $X$ to the original. We must add that $W$ may represent additive noise but also modeling errors. Indeed since $\G(Z)$ is usually some finite dimensional neural network it cannot express exactly the desired random vector $X$ and therefore $\T(X,\alpha)$ does not represent exactly the observations $Y$ even if they are noiseless. It is this representation error that we also model as additive ``noise''.

Of course the goal is from $Y$ to obtain an estimate $\hat{X}$. However such an estimate would require the joint density $\f(Y,X)$ which is not available. We propose instead a two-step alternative: In the first step we estimate $\hat{Z}$, that is, the proper input to the generator that gives rise to the observations $Y$; in the second step we \textit{define} $\hat{X}=\G(\hat{Z})$. Indeed this process makes sense. If the generator model represents very closely the statistical behavior of the random vector $X$, this means that, to each realization of $X$ there exists a realization of $Z$ such that $X=\G(Z)$. Consequently it is the input to the generator we are seeking first and then we use it to generate the corresponding output.

It is not difficult to verify (see \cite{basioti} for details) that the joint density of $Y$ and $Z$ is given by
$$
\f(Y,Z|\alpha,\beta)=\g\left(Y-\T\big(\G(Z),\alpha\big)|\beta\right)\h(Z)
$$
where the combination $\{\alpha,\beta\}$ replaces the parameter vector $\gamma$ of our general theory. We can now apply \eqref{eq:optim} and this yields
\begin{equation}
\hat{Z}=\text{arg}\max_Z\left\{\max_{\alpha,\beta}\g\left(Y-\T\big(\G(Z),\alpha\big)|\beta\right)\right\}\h(Z).
\label{eq:mbifla1}
\end{equation}
If, additionally, we assume that the additive noise is Gaussian with mean 0 and covariance matrix $\beta^2 I$ where $I$ is the identity matrix, i.e.~$\g(W|\beta)=\mathcal{N}(0,\beta^2I)$, and that the input density is also Gaussian with mean 0 and covariance $I$, that is, $\h(Z)=\mathcal{N}(0,I)$ (we recall that $\h(Z)$ can be selected by us and the most common forms are i.i.d.~uniform or i.i.d.~standard normal), then the previous estimate becomes
\begin{equation}
\hat{Z}=\text{arg}\min_Z\left\{ N \log (\min_\alpha \| Y - \T(\G(Z),\alpha)\|^2) + \|Z\|^2\right\},
\label{eq:mbifla2}
\end{equation}
where $N$ denotes the size of $Y$ and $W$ and where the previous expression is obtained from \eqref{eq:mbifla1} by finding explicitly the optimum $\beta$ and substituting.

If we are going to employ some version of steepest decent to iteratively solve \eqref{eq:mbifla2} then we can see that the previous problem is equivalent to the solution of the following minimization problem
\begin{equation}
\min_{Z,\alpha}\left\{ N \log (\| Y - \T(\G(Z),\alpha)\|^2) + \|Z\|^2\right\},
\label{eq:mbifla3}
\end{equation}
where each iteration of the steepest descent will provide parallel estimates $\{\hat{Z}_t,\hat{\alpha}_t\}$ of $Z$ and $\alpha$. Of course once the iterations converge then we only need the final estimate for $Z$.

When we are going to apply this general theory to the problem of interest we will specify, more precisely, the transformation $\T(\cdot)$. But even at this point we can claim a number of interesting characteristics that our proposed method is going to enjoy: 
\begin{itemize}
\item It is the result of a rigorous mathematical analysis based on the statistical estimation theory. 
\item It does not contain any unknown weights that require fine-tuning with the help of training data.
\item It can accommodate transformation $\T(X,\alpha)$ that contain unknown parameters $\alpha$ that \textit{may change with every realization} and therefore it is not possible to fine-tune them before hand. The last characteristic is truly exceptional and not encountered in any other generative model based method that offers a solution to the problem of interest.
\end{itemize}
Let us now apply this general idea to various versions of the de-quantization problem.

\section{Application to De-Quantization}
Suppose that we have an image $X=\{x_1,\ldots,x_n\}$ where $\{x_i\}$ are the image pixels. If the image is colored then we know that each $x_i$ is a 3-D vector containing the corresponding three chromatic components. We assume that each chromatic component takes value in the interval $[0,1]$. If our quantization method produces $M$ colors $Q=\{q_1,\ldots,q_M\}$ with each $q_i$ being a 3-D vector containing its corresponding chromatic components then the transformation $\T(X)$ is applied to each pixel $x_i$ separately and we have $\T(X)=\{\T(x_1),\ldots,\T(x_n)\}$ with $\T(x)$ defined as follows
\begin{equation}
\T(x)=\text{arg}\min_{q\in Q}\{\|x-q\|\}.
\label{eq:mbafla1}
\end{equation}
In other words, for each pixel $x$, we select the color from $Q$ which is closest (in the Euclidean distance sense) to its 3-D chromatic vector. Unfortunately transformations of the form of \eqref{eq:mbafla1} are not differentiable, a property which is necessary when we employ steepest descent type iterative solvers for \eqref{eq:mbifla3}. Even though \eqref{eq:mbafla1} is the quantizer used to produce the actual data $Y$, for the estimation of $Z$ we need to replace it with a function which is differentiable. We propose the following softmax alternative
\begin{equation}
\tilde{\T}(x,k)=\frac{\sum_{i=1}^M q_i e^{-k\|x-q_i\|^2}}{\sum_{i=1}^M e^{-k\|x-q_i\|^2}},~k>0.
\label{eq:mbafla2}
\end{equation}
Indeed we observe that as $k\to\infty$ we have that $\tilde{\T}(x,k)\to\T(x)$. In fact it is better if we let $k$ to be a parameter to be optimized instead of fixing it to some arbitrary value. This means that $k$ will be a component of $\alpha$ in \eqref{eq:mbifla3}.

\begin{figure}[!t]
\vskip0.2cm
\centering
\includegraphics[width=2.0in]{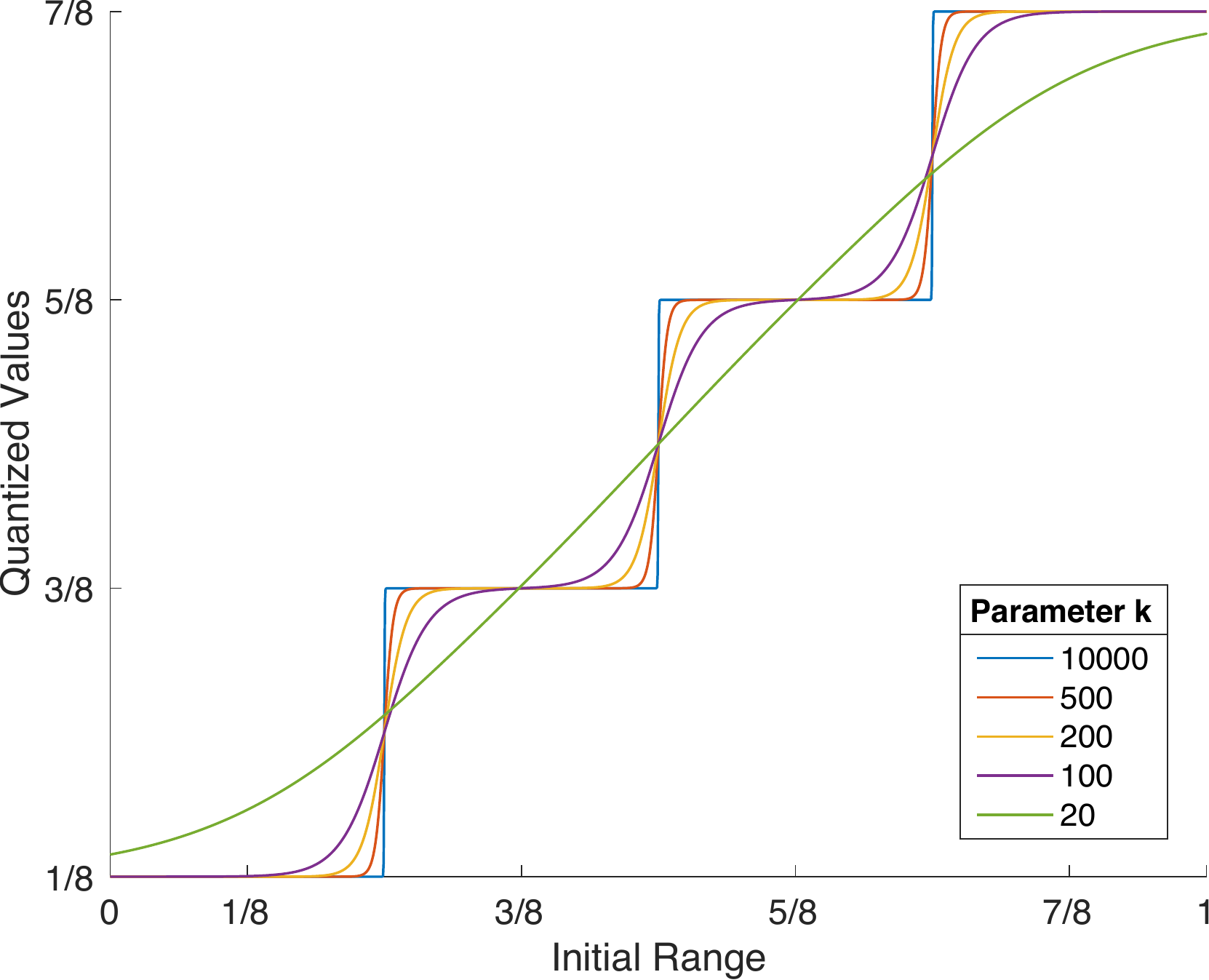}
\vskip-0.2cm
\caption{Softmax approximation $\tilde{\T}(r,k)$ defined in \eqref{eq:mbafla4} of the uniform quantization function $\T(r)$ defined in \eqref{eq:mbafla3} for $m=4$ and different values of the parameter $k$.}
\label{fig:1}
\vskip-0.4cm
\end{figure}

If we adopt a more classical quantization scheme as for example uniform quantization and the same number of levels $m$ in each chromatic component, then the previous transformation simplifies and instead of applying it per pixel using the 3-D chromatic vectors we can use the same idea and apply it in each pixel and each chromatic component separately. This means that if $X=\{x_1,\ldots,x_n\}$ and each pixel $x_i=\{r_i,g_i,b_i\}$ where $r,g,b$ denote the red, green, blue components then $\T(X)=\{\T(r_1),\T(g_1),\T(b_1),\cdots,\T(r_n),\T(g_n),\T(b_n)\}$ with
\begin{equation}
\T(r)=\sum_{i=1}^m\frac{i-0.5}{m}\mathds{1}_{\{i-1<r\leq i\}},
\label{eq:mbafla3}
\end{equation}
where $\mathds{1}_{A}$ denotes the indicator function of the set $A$. We have a similar definition for $\T(g)$ and $\T(b)$. It is also clear that the total number of different colors is 
$M=m\times m\times m=m^3$. For the solution of \eqref{eq:mbifla3} we need to replace the transformation with a differentiable alternative and we can use the softmax version
\begin{equation}
\tilde{\T}(r,k)=\frac{\sum_{i=1}^m\frac{i-0.5}{m}e^{-k(r-i-0.5)^2}}{\sum_{i=1}^m e^{-k(r-i-0.5)^2}},
\label{eq:mbafla4}
\end{equation}
where, again, as we can see from Fig.\,\ref{fig:1}, we have $\tilde{\T}(r,k)\to\T(r)$ as $k\to\infty$.

Let us now present different de-quantization challenges using the popular CelebA \cite{liu} dataset which contains 202,599 RGB images that were cropped and resized to $64\times64\times3$ and then separated into two sets 202,499 for training and 100 for testing. 
The proposed methodology will be tested against the 100 testing images. 

Since our method requires a generative model we will use the training set in order to design it and it will be the same for all de-quantization problem we present next. We employ the adversarial approach and in particular the progressive GAN method developed in \cite{karras}. We adopt the following configuration: For the generator we use input of size 512 and i.i.d.~Gaussian elements $\mathcal{N}(0,1)$ as discussed in Section\,\ref{sec:3}. We have five layers and each layer consists of two convolutions with two kernels $3\times3$ except the first layer that has one $4\times4$ and one $3\times3$ kernel and the last that has two $3\times3$ and one $1\times1$ kernel resulting in an output of $64\times64\times3$. After each convolution, a leaky ReLU is applied except for the last $1\times1$ convolution where no nonlinear function is employed. The intermediate layers also involve an upsampling operation. For the discriminator we use input $64\times64\times3$ and six layers in total. The first five layers have two convolutions with two $3\times3$ kernels except for the first layer which has an additional $1\times1$ layer and the last layer which has a $3\times3$ and a $4\times4$ kernel. After each convolution, we apply a leaky ReLU except for the last $4\times4$ kernel where no nonlinearity is used. In the intermediate layers, we apply downsampling except for the last layer. Finally, we employ a fully connected part that provides the scalar output of the discriminator. We recall that in our approach the discriminator is not used since the generator along with the input density are supposed to completely capture the statistical behavior of the dataset.

Before we proceed with our simulations we would like to discuss a slight variation of our main de-quantization method based on an idea proposed in \cite{yeh2} as a simpler alternative to \eqref{eq:mbifla3}. Instead of using the transformation $\T(X)$, in \cite{yeh2} they replace it with the identity, namely $\T(X)=X$. This idea used in our approach consists in replacing \eqref{eq:mbifla3} with the simpler optimization problem
\begin{equation}
\min_{Z}\left\{ N \log (\| Y - \G(Z)\|^2) + \|Z\|^2\right\},
\label{eq:mbifla33}
\end{equation}
where no quantization scheme is applied on the output of the generator and it is simply contained in the observations $Y$.
As we will have the chance to confirm in the simulations that follow, this simplification has quite satisfactory performance if the number $M$ of colors after quantization is relatively high. However, when $M$ is small it may fail miserably.

\subsection{Uniform quantization on all chromatic components}
Let us start with the simplest version of the problem where we quantize the chromatic components uniformly as described in  \eqref{eq:mbafla3}. We are going to see the performance for different values of the number of levels $m$. For quantization we use the transformation defined in \eqref{eq:mbafla3} but for de-quantization in \eqref{eq:mbifla3} we employ the approximation defined in \eqref{eq:mbafla4}. 
\begin{figure}[!h]
\vskip0.2cm
\centerline{\includegraphics[width=0.98\hsize]{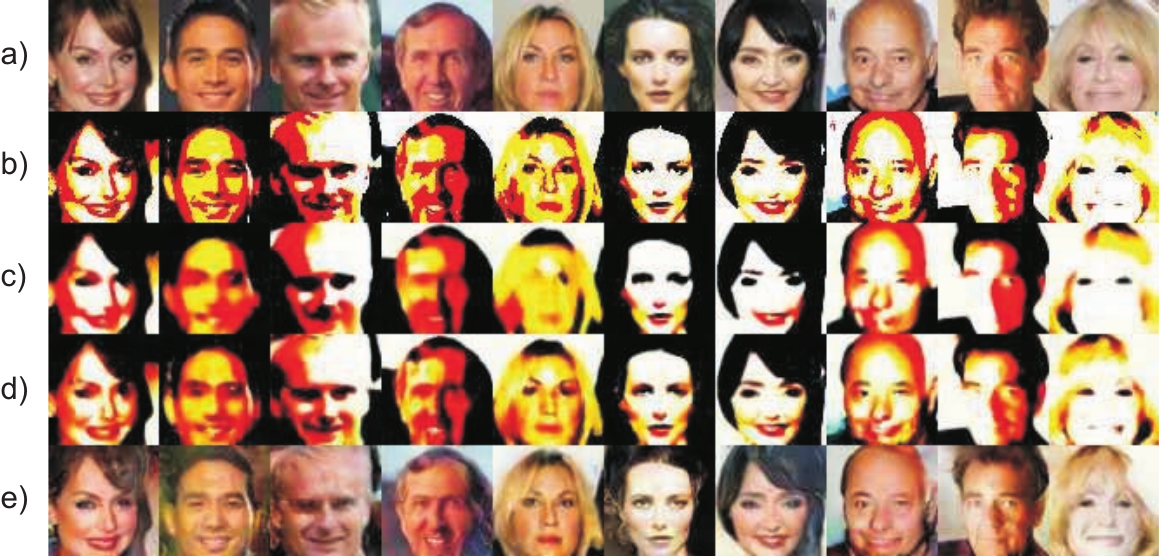}}
\vskip-0.1cm
\centerline{\small $m=2$, 8-color quantization}
\vskip0.2cm
\centering
\includegraphics[width=0.98\hsize]{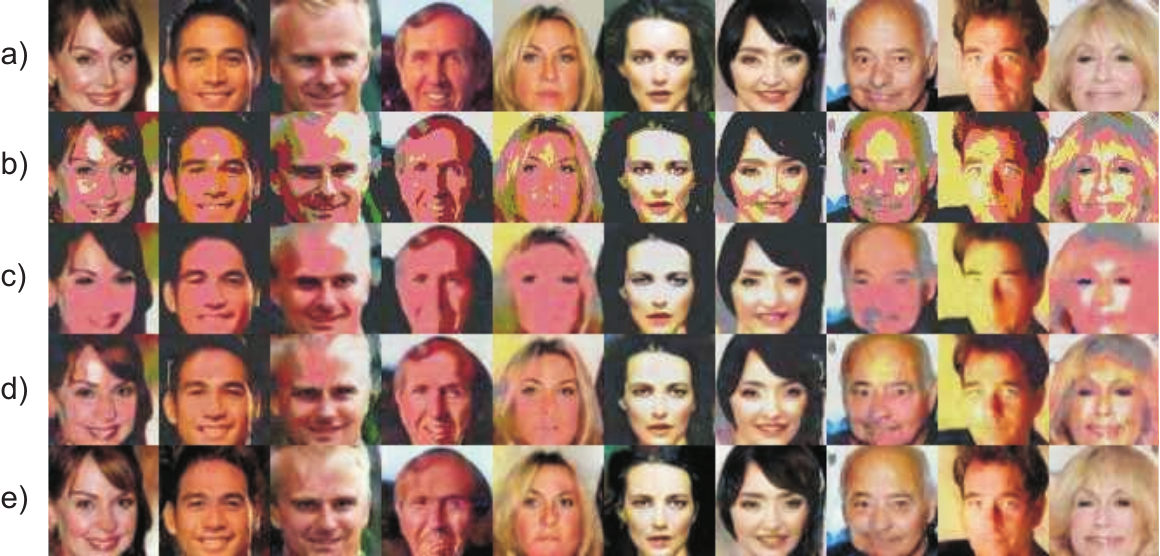}
\vskip-0.1cm
\centerline{\small $m=3$, 27-color quantization}
\vskip0.2cm
\centering
\includegraphics[width=0.98\hsize]{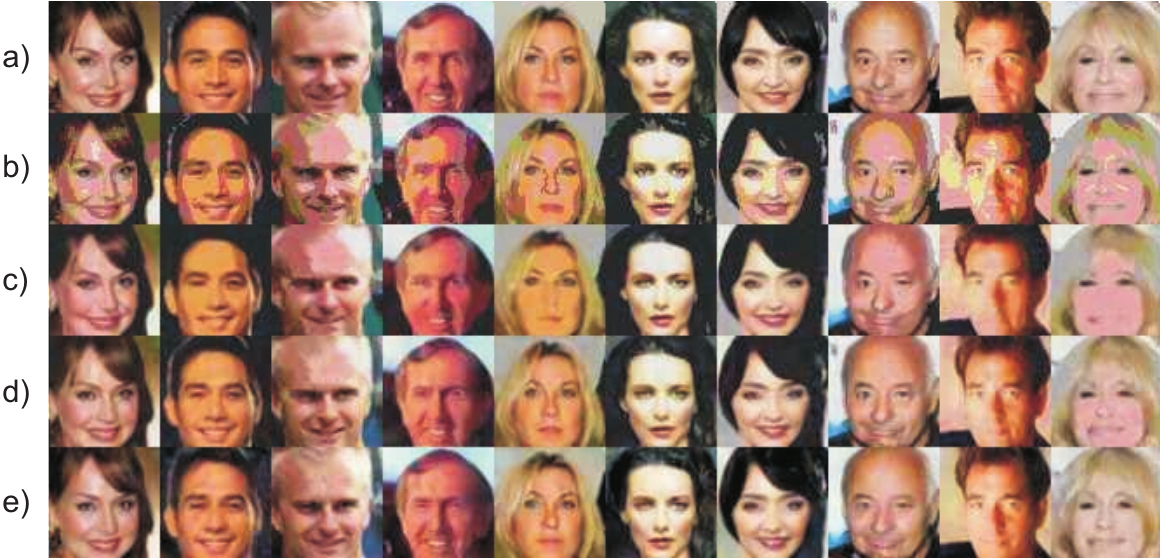}
\vskip-0.1cm
\centerline{\small $m=4$, 64-color quantization}
\vskip0.2cm
\centering
\includegraphics[width=0.98\hsize]{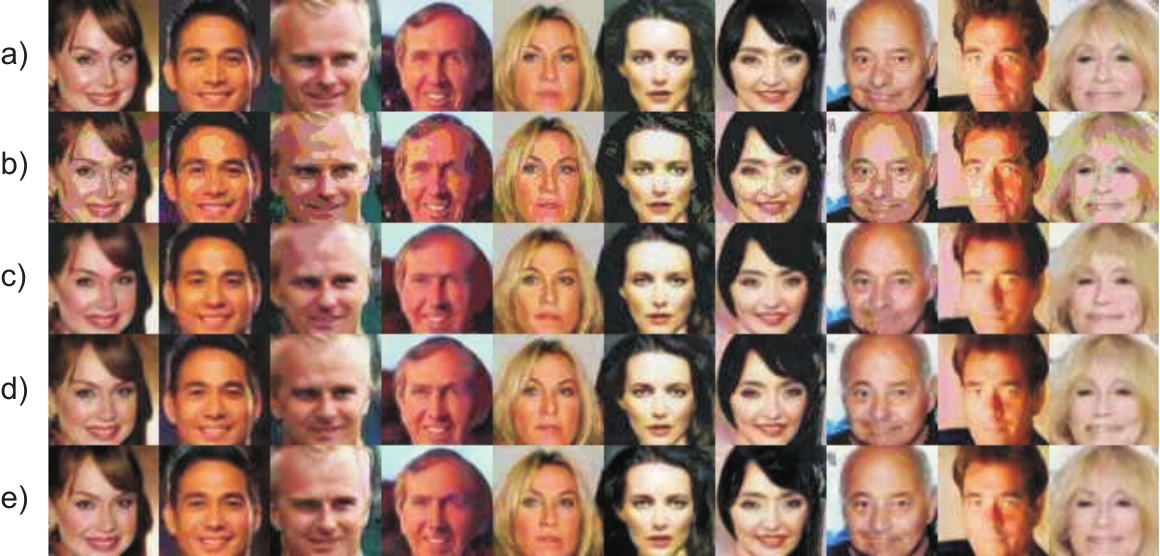}
\vskip-0.1cm
\centerline{\small $m=5$, 125-color quantization}
\vskip-0.2cm
\caption{De-quantization results with $m=2,3,4,5$ levels per chromatic component. Row a)~Original; b)~Quantized; c)~\cite{yeh2}, identity transform; d)~Proposed, identity transform; e)~Proposed, softmax transform.}
\label{fig:2}
\vskip-0.5cm
\end{figure}
If $n$ is the number of pixels then in this case $N=3n$ since each pixel contains three chromatic components.

In all competing methods, in this and all 
subsequent simulations, we apply the momentum gradient descent \cite{qian} with normalized gradients where the momentum hyperparameter is set to 0.999, the learning rate to 0.1 and each algorithm
 runs for 200,000 iterations.

As we can see in Fig.\,\ref{fig:2} when the number $m$ of quantization levels is small and of the order of 2 or 3 then the proposed method when we employ the softmax transformation defined in \eqref{eq:mbafla4} has a notably superior performance to the other two alternatives. When however $m$ increases to 4 and 5 resulting in a high number of overall number of colors then this difference becomes less pronounced. 
What we can observe from 
\begin{table}[!h]
\renewcommand{\arraystretch}{1.3}
\caption{Reconstruction Errors and PSNRs}
\label{table_example}
\centering
\begin{tabular}{|c||c||c||c||c||c||c|}
\hline
          & \multicolumn{2}{c||}{Yeh} & \multicolumn{2}{c||}{Prop. Identity} & \multicolumn{2}{c|}{Prop. Softmax} \\ \hline
 $M=m^3$             & Error      & PSNR        & Error           & PSNR            & Error          & PSNR            \\ \hline
8   & 0.071   &  13.105   & 0.046          & 13.549         & 0.012         & 19.496         \\ \hline
27  &  0.008   & 20.500   & 0.005          & 23.091         & 0.004         & 23.774         \\ \hline
64  &  0.004  &   23.221 & 0.003          & 25.369         & 0.003         & 25.848         \\ \hline
125  &   0.003  &25.008& 0.002  & 26.743  & 0.002 & 26.746 \\ \hline
\end{tabular}
\label{tab:1}
\end{table}
the figure it is also corroborated by Table\,\ref{tab:1} where we present the reconstruction errors and the peek to SNR index. As we see our proposed method based on the softmax approximation of the quantizer for $m=2 (M=8)$ has 7 and 4 times smaller error compared to the method in \cite{yeh} and the simplified version of our approach. These factors when $m=3 (M=27)$ become 2 and 1.2 respectively.

\subsection{De-quantization and colorization}
Let us now combine de-quantization with the colorization problem. In particular we assume that first we apply a linear transformation where from the RGB representation we obtain a grayscale intensity equivalent and then we apply a scalar uniform quantization scheme on the intensity of each pixel. Let us make explicit the resulting transformation. First we recognize three matrices $\T_{\rm R},\T_{\rm G},\T_{\rm B}$ where each one isolates the corresponding RGB chromatic component. In particular $\T_{\rm R}X$ isolates the red component of each pixel. The components are then combined as $\I=a_{\rm R}\T_{\rm R}X+a_{\rm G}\T_{\rm G}X+a_{\rm B}\T_{\rm B}X$ to form the intensity vector $\I$. We recall that the combination coefficients $a_{\rm R}=0.2126,a_{\rm G}=0.7152,a_{\rm B}=0.0722$ are considered as the ideal values to transform an RGB to grayscale. Once we have the intensity $\I$ we quantize each pixel intensity uniformly using \eqref{eq:mbafla3} with $\I$ replacing $r$ and this constitutes the monochromatic quantized vector $Y$. The goal, as before, is from the quantized grayscale image $Y$ \textit{to recover the original RGB image $X$}.

The next simulation focuses on exactly this problem. Specifically we intend to adopt a rather severe quantization policy with only two levels 0 and 1. If $\I_i$ denotes the intensity of the $i$-th pixel then this is quantized to the value 
$\mathds{1}_{\{\I_i\geq0.5\}}$. In other words we use as threshold the middle of the intensity range generating a uniform two-level quantization. Of course if we would like to use this scheme for de-quantization we need to replace the indicator function with a differentiable alternative. 
As such we can use the sigmoid $\frac{1}{1+e^{-k(\I_i-0.5)}}$ where again as $k\to\infty$ we approach the ideal quantizer. Of course we will consider $k$ to be a parameter to be optimized through the optimization problem defined in \eqref{eq:mbifla3} in order to find the sigmoid approximation that best fits the data. In this case if $n$ is the number of pixels then $N=n$ since for each pixel we have only one intensity value.

\begin{figure}[!h]
\vskip-0.3cm
\centering
\includegraphics[width=0.98\hsize]{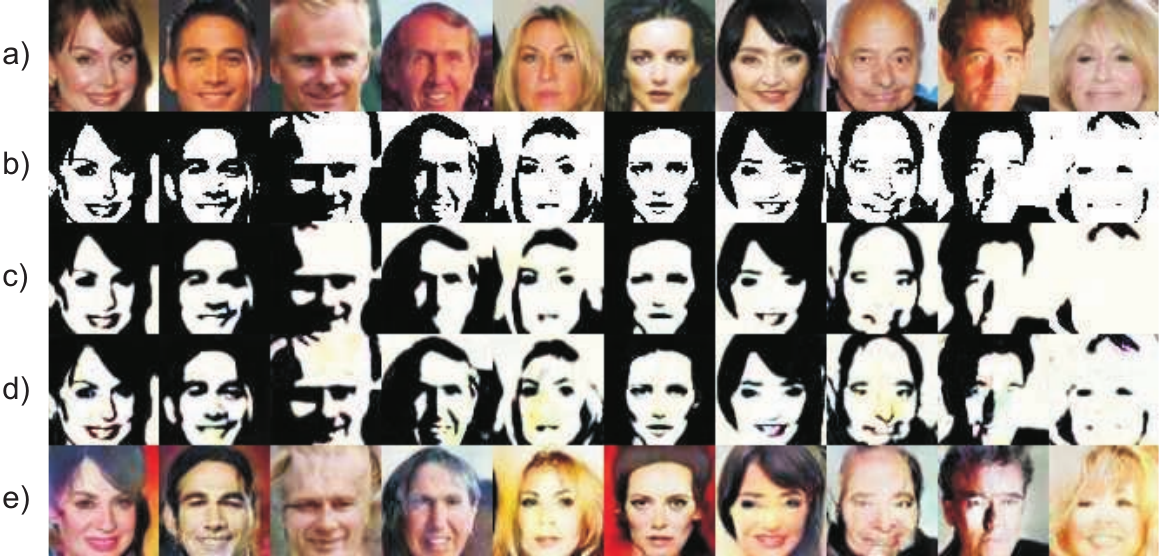}
\vskip-0.1cm
\caption{De-quantization and colorization of 2-level gray images. Row a)~Original; b)~Black and white version; c)~\cite{yeh2}, identity transform; d)~Proposed, identity transform; e)~Proposed, sigmoid transform.}
\label{fig:3}
\vskip-0.2cm
\end{figure}
As we can see from Fig.\,\ref{fig:3} only our full method with the quantization scheme approximated by the sigmoid obtains extremely satisfactory results. We not only de-quantize but we also colorize successfully recovering correctly the chromatic components. This is rather extraordinary considering the very limited information contained in the quantized black and white data. Apparently, the generative model contains sufficient prior information which is capable, at least for this class of images, to fill the gaps and successfully restore the original images.

We must stress that in this example the transformation leading to the quantized values was considered completely known. In other words the combination coefficients $a_{\rm R}=0.2126,a_{\rm G}=0.7152,a_{\rm B}=0.0722$ and the quantizer threshold 0.5 where used in the restoration process.

\subsection{De-quantization, colorization and parameter estimation}
Let us now consider again the transformation from RGB to grayscale using the same combination we mentioned in our previous example. However, in this example the two-level quantization threshold is no longer set to 0.5 but to $\delta$ with the latter being \textit{unknown} during the restoration process. Such a situation arises when for instance for the two level quantization we use the Otsu thresholding scheme where the threshold $\delta$ is image (realization) dependent and computed with the method proposed in \cite{otsu}. Consequently, at each pixel, quantization is of the form $\mathds{1}_{\{\I_i\geq\delta\}}$ where $\delta\in(0,1)$ and, as mentioned, for the de-quantization and colorization procedure $\delta$ is an unknown parameter. For restoration we need to approximate the index function with a differentiable alternative. We propose the sigmoid of the form $\frac{1}{1+e^{-k(\I_i-\delta)}}$ which is centered around the unknown threshold $\delta$. Both parameters $k,\delta$ are considered unknown and the optimization problem in \eqref{eq:mbifla3} must include a minimization with respect to $Z,k,\delta$. As in the previous experiment $N=n$ since, again, each pixel has one intensity value.

\begin{figure}[!h]
\centering
\includegraphics[width=0.98\hsize]{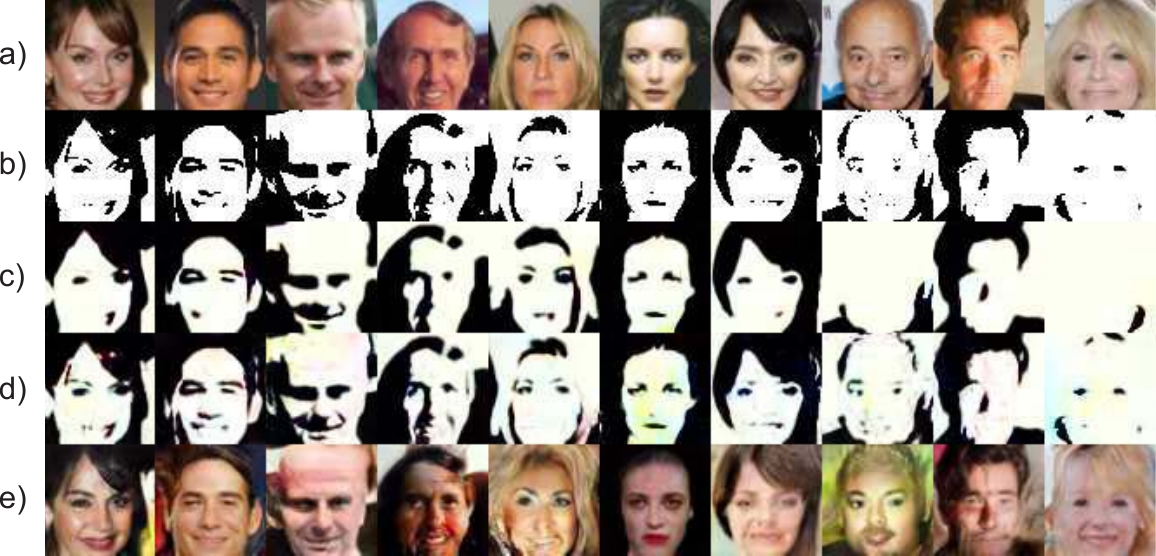}
\vskip-0.2cm
\caption{De-quantization and colorization of 2-level gray images. Row a)~Original; b)~Black and white version; c)~\cite{yeh2}, identity transform; d)~Proposed, identity transform; e) Proposed, sigmoid transform and threshold estimation.}
\label{fig:4}
\vskip-0.2cm
\end{figure}
Fig.\,\ref{fig:4} captures the corresponding results when $\delta$ is set to 0.4. In the method of \cite{yeh2} and our version with the simplified optimization problem where the transformation is replaced by the identity the exact knowledge of the quantizer is not required. In the same figure, the last row contains our results when the threshold is unknown 
\begin{figure}[!h]
\centering
\includegraphics[width=0.98\hsize]{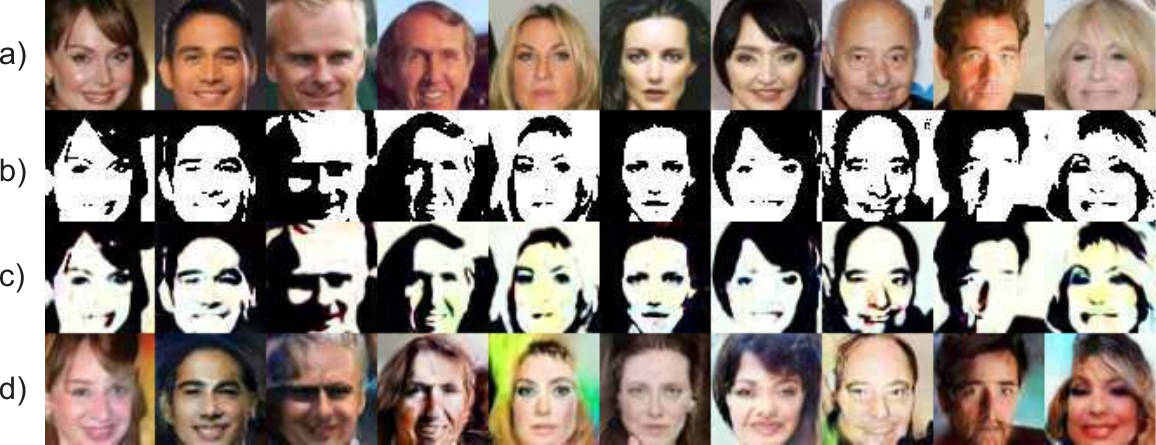}
\vskip-0.2cm
\caption{De-quantization and colorization of 2-level gray images. Row a)~Original; b)~Black and white version; c)~Proposed, identity transform; d)~Proposed, sigmoid transform and threshold estimation.}
\label{fig:6}
\end{figure}
and we estimate it along with $k$ and $Z$. Fig.\,\ref{fig:6} contains a similar experiment only now the threshold $\delta$ is set by the Otsu thresholding scheme.
In this case the method in \cite{yeh2} cannot be applied because it is impossible to fine-tune the weight of the regularizer term since it requires a constant transformation $\T(\cdot)$. As we mentioned, this is not the case here because the transformation contains parameters that are realization dependent (change with every image). For this reason we included only our simplified method with the results appearing in the third row, while we reserved the last row for the full method.

\begin{figure}[!h]
\centering
\includegraphics[width=2.2in]{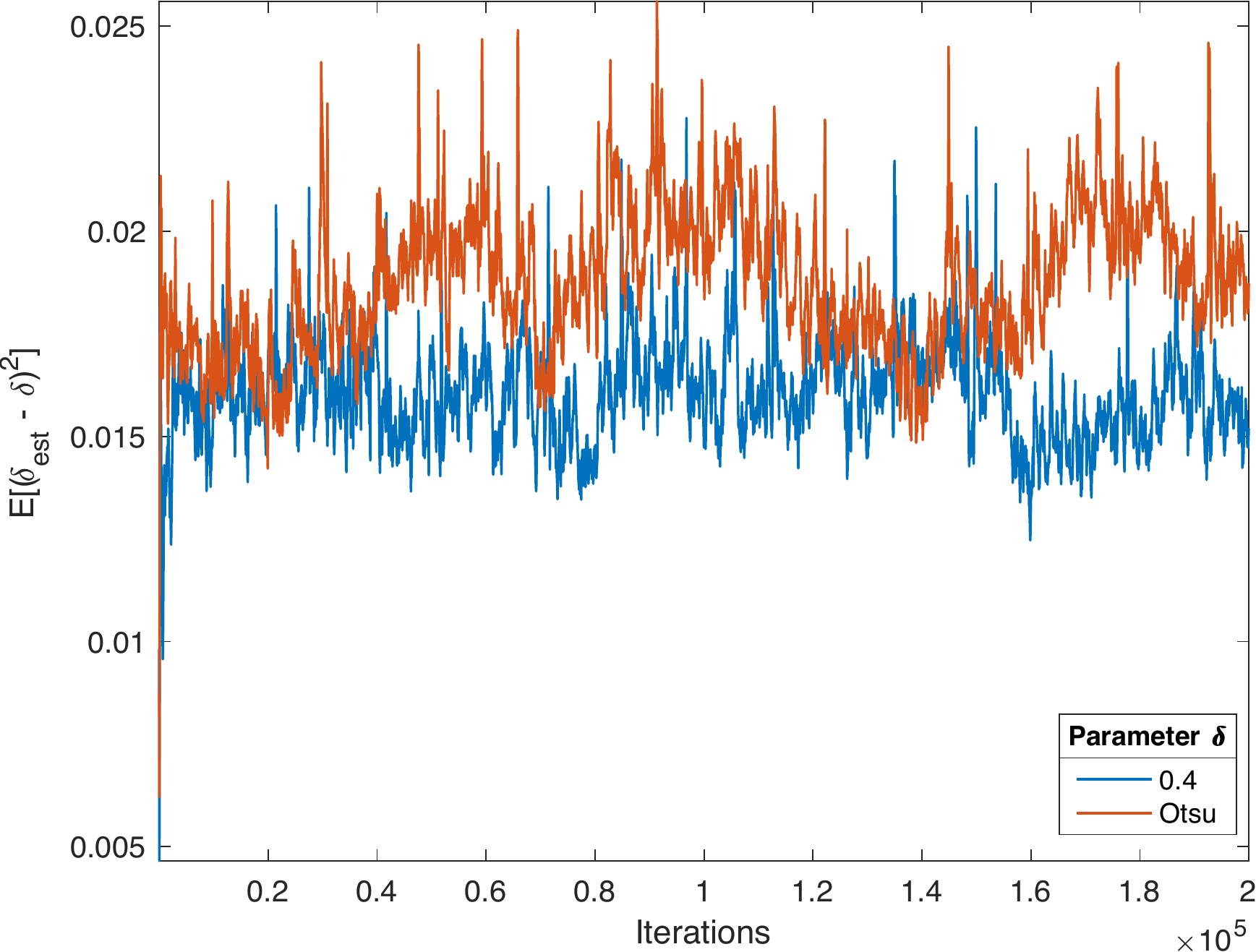}
\vskip-0.3cm
\caption{Estimation error variance of $\delta$ as a function of the number of iterations.}
\label{fig:5}
\vskip-0.2cm
\end{figure}
Fig.\,\ref{fig:5} contains the evolution of the estimation error variance of $\delta$ with the number of iterations. We have included the two cases we presented before namely $\delta=0.4$ (blue) and $\delta$ set by the Otsu 
thresholding mechanism (red). As we can see our parallel $\delta$ estimator provides relatively reasonable estimates considering the severe deformation the original image has been subjected to.
\begin{table}[!h]
\vskip-0.3cm
\renewcommand{\arraystretch}{1.3}
\caption{Reconstruction Errors and PSNRs}
\label{table_example}
\vskip-0.2cm
\centering
\begin{tabular}{|c||c||c||c||c||c||c|}
\hline
          & \multicolumn{2}{c||}{Yeh} & \multicolumn{2}{c||}{Prop. Identity} & \multicolumn{2}{c|}{Prop. Softmax} \\ \hline
 $M=2$             & Error      & PSNR        & Error           & PSNR            & Error          & PSNR            \\ \hline
$\delta = 0.5$  & 0.092     & 11.770     & 0.076          & 11.474        & 0.029         & 15.841         \\ \hline
$\delta = 0.4$  &  0.112   & 11.693   & 0.087          & 10.956         & 0.029         & 15.750         \\ \hline
$\delta = \text{Otsu}$  &  --  &   -- & 0.085          & 11.079         & 0.033         & 15.744         \\ \hline
\end{tabular}
\label{tab:2}
\vskip-0.1cm
\end{table}
The satisfactory performance of our full method is also depicted in Table\,\ref{tab:2} where the indices suggest that it enjoys significantly better performance compared to its competitors.

\section{Conclusion}
We have presented a de-quantization method based on a generative modeling of the ideal image. The proposed technique requires the solution of an optimization problem that was derived through rigorous mathematical analysis of the de-quantization problem and using ideas borrowed from statistical estimation theory. Our optimization problem is completely specified without any unknown weights attached to a regularizer term that require fine-tuning before hand. We must also mention that our processing scheme is capable of treating problems with unknown parameters as for example in two level quantization when the quantization threshold is unknown. In such cases it simultaneously estimates the parameters and restores the original image very successfully.

\section*{Acknowledgement}
This work was supported by the US National Science Foundation under Grant CIF\,1513373, through Rutgers University.


\begin{thebibliography}{99}

\bibitem{pollard}
D.~Pollard, ``Quantization and the method of k-means,''  \emph{ IEEE Transactions on Information theory}, vol.~28, no.~2, pp.~199-205, 1982.


\bibitem{afonso}
M.~V. Afonso, and J.~M. Bioucas-Dias and M.~A. T. Figueiredo, ``An augmented Lagrangian approach to the constrained optimization formulation of imaging inverse problems,''  \emph{IEEE Trans. Image Proc.}, vol.~20, no.~3, pp.~681-695, 2010.

\bibitem{hu}
Y. Hu, and D. Zhang, and J. Ye, and X. Li, and X. He, ``Fast and accurate matrix completion via truncated nuclear norm regularization,'' \emph{IEEE Trans. Patt. Anal. Mach. Intel.}, vol.~35, no.~9, pp.~2117-2130, 2012.

\bibitem{yang}
J. Yang, and J. Wright, and T. S. Huang, and Y. Ma, ``Image super-resolution via sparse representation,'' \emph{IEEE Trans. Image Proc.}, vol.~19, no.~11, pp.~2861-2873, 2010.

\bibitem{bora}
A. Bora, and A. Jalal, and E. Price, and A. G. Dimakis, ``Compressed sensing using generative models,'' \emph{arXiv: 1703.03208}, 2017.

\bibitem{yeh}
R. A. Yeh, and C. Chen, and T. Yian Lim, and A. G. Schwing, and M. Hasegawa-Johnson, and M. N. Do, ``Semantic image inpainting with deep generative models,''  \emph{Proc. IEEE Conf. Comp. Vision Patt. Recogn.}, pp.~5485-5493, 2017.

\bibitem{yeh2}
R. A. Yeh, and T. Y. Lim, and C. Chen, and A. G. Schwing, and M. Hasegawa-Johnson, and M. N. Do, ``Image restoration with deep generative models,'' \emph{IEEE Intern. Conf. Acoust. Speech Sig. Proc.}, pp. 6772-6776, 2018.

\newpage
\bibitem{basioti}
K.~Basioti and G.~V. Moustakides, ``Image restoration from parametric transformations
using generative models,'' \emph{arXiv: 2005.14036}, 2020.

\bibitem{veeravalli}
P. Moulin and V.~V.~Veeravalli, {\em Statistical Inference for Engineers and Data Scientists}, Cambridge University Press, 2019.

\bibitem{goodfellow}
I. Goodfellow, and J. Pouget-Abadie, and M. Mirza, and B. Xu, and D. Warde-Farley, and S. Ozair, and A. Courville, and Y. Bengio, ``Generative adversarial nets,'' \emph{Advances in Neural Information Processing Systems}, pp.~2672-2680, 2014.

\bibitem{karras}
T. Karras, and T. Aila, and S. Laine, and J. Lehtinen, ``Progressive growing of GANs for improved quality, stability, and variation,''  \emph{arXiv: 1710.10196}, 2017.

\bibitem{basioti2}
K. Basioti, and G. V. Moustakides, ``Designing GANs: A likelihood ratio approach,'' \emph{arXiv: 2002.00865}, 2020.

\bibitem{basioti3}
K. Basioti, and G. V. Moustakides, and E. Z. Psarakis,  ``Maximal correlation: An alternative criterion for training generative Networks,'' \emph{Proc. 24th Europ. Conf. Artif. Intel.}, 2020.

\bibitem{dziugaite}
G. K. Dziugaite, and D. M. Roy, and Z. Ghahramani, ``Training generative neural networks via maximum mean discrepancy optimization,'' \emph{arXiv: 1505.03906}, 2015.

\bibitem{doersch}
C. Doersch, ``Tutorial on variational autoencoders,''  \emph{arXiv: 1606.05908}, 2016.

\bibitem{liu}
Z. Liu, and P. Luo, and X. Wang, and X. Tang, ``Deep learning face attributes in the wild,'' \emph{Proc. IEEE Intern. Conf. Comp. Vision}, pp.~3730-3738, 2015.

\bibitem{qian}
N. Qian, ``On the momentum term in gradient descent learning algorithms,'' \emph{Neural Networks}, vol.~12, no.~1, pp.~145-151, 1999.

\bibitem{otsu}
N.~Otsu, ``A threshold selection method from gray-level histograms,'' \emph{IEEE Trans. Syst. Man and Cyber.}, vol.~SMC-9, no.~1, pp.~62-66, 1979.

\end{thebibliography}
\end{document}